\documentclass{article}
\usepackage{PRIMEarxiv}
\usepackage{float}
\usepackage[utf8]{inputenc} 
\usepackage[T1]{fontenc}    
\usepackage{hyperref}       
\usepackage{url}            
\usepackage{booktabs}       
\usepackage{amsfonts}       
\usepackage{nicefrac}       
\usepackage{microtype}      
\usepackage{lipsum}
\usepackage{fancyhdr}       
\usepackage{graphicx}       
\graphicspath{{media/}}     
\usepackage{soul,color}
\usepackage{multirow}
\usepackage{amsmath}
\usepackage{amsbsy}
\usepackage{amssymb}

\usepackage{bm}
\title{Disturbing Image Detection Using LMM-Elicited Emotion Embeddings
\thanks{\textit{\underline{Citation}}: 
\textbf{M. Tzelepi, V. Mezaris, "Disturbing Image Detection Using LMM-Elicited Emotion Embeddings", Proc. Integrating Image Processing with Large-Scale Language/Vision Models for Advanced Visual Understanding (LVLM) Workshop at IEEE Int. Conf. on Image Processing (ICIP 2024), Abu Dhabi, United Arab Emirates, Oct.2024, accepted for publication.}}
\thanks{(c) IEEE. This is the authors' accepted version. The final IEEE-published version can be found in IEEE Xplore.  }
}

\author{
  Maria Tzelepi and Vasileios Mezaris\\ 
  Information Technologies Institute (ITI) \\
  Centre of Research and Technology Hellas (CERTH) \\
  Thessaloniki, Greece\\
  \texttt{\{mtzelepi,bmezaris\}@iti.gr} \\
}

\begin{document}
\maketitle
%
%
%
\begin{abstract}
In this paper we deal with the task of Disturbing Image Detection (DID), exploiting knowledge encoded in Large Multimodal Models (LMMs). Specifically, we propose to exploit LMM knowledge in a two-fold manner: first by extracting generic semantic descriptions, and second by extracting elicited emotions. Subsequently, we use the CLIP's text encoder in order to obtain the text embeddings of both the generic semantic descriptions and LMM-elicited emotions. Finally, we use the aforementioned text embeddings along with the corresponding CLIP's image embeddings for performing the DID task. The proposed method significantly improves the baseline classification accuracy, achieving state-of-the-art performance on the augmented Disturbing Image Detection dataset.
\end{abstract}

%
%
\section{Introduction}
Disturbing Image Detection (DID) refers to the task of detecting content in images that can cause trauma to the viewers \cite{zampoglou2016web,sarridis2024mitigating}. It may include images that depict violence, pornography, animal cruelty, disasters. 
Such content elicits anxiety or/and fear to viewers, while it is noted that viewers who are systematically exposed to such content are at risk of experiencing severe psychological distress \cite{dubberley2015making}, rendering the DID a task of significant importance. 

The literature on the specific task of the DID is somewhat limited, primarily due to the challenging nature of creating datasets, restricting, in turn, the generalization ability of the models trained with few images \cite{larocque2021gore}. The DID dataset, presented in \cite{zampoglou2016web}, is the largest one, considering the DID task, being however a small dataset that consists of 5,401 images. In \cite{zampoglou2016web}, an SVM classifier was trained for detecting disturbing images, using features extracted from a convolutional neural network. Subsequently, aiming to tackle the issue of limited training samples, a framework that exploits large-scale multimedia datasets so as to automatically extend initial training datasets with hard examples, was applied to the abovementioned dataset in \cite{sarridis2022leveraging}. In the latter work, an EfficientNet-b1 \cite{tan2019efficientnet} was trained on the augmented dataset for addressing the DID task. Other relevant but distinct works focus on Not Safe for Work (NSFW) detection \cite{gangwar2021attm}, while there are also works that focus on violence detection \cite{mu2016violent}.

In this paper, we propose to address the DID task by exploiting knowledge encoded in Large Language Models (LLMs) \cite{zhao2023survey}, and particularly Large Multimodal Models (LMMs) (also known as Multimodal Large Language Models) \cite{yin2023survey}. LLMs/LMMs have demonstrated exceptional performance in several downstream vision recognition tasks over the recent few years, gradually displacing former deep learning approaches \cite{zhang2024mm}. To this end, we propose to use MiniGPT-4 \cite{zhu2023minigpt} in order to extract semantic descriptions for the images of the dataset. Apart from the aforementioned generic descriptions, we propose to extract responses linked with a complementary task, i.e., emotion recognition. That is, we argue that we can advance the performance in the DID task by also extracting LMM-elicited emotions for each image of the dataset. Then, we use a popular Vision Language Model (VLM), i.e., CLIP \cite{radford2021learning}, in order to leverage the aforementioned knowledge into our classification task. Specifically, we extract the CLIP text embeddings of the generic semantic descriptions and the elicited emotions. We finally use the aforementioned text embeddings along with the corresponding CLIP image embeddings for performing the DID task. Note that obtaining the CLIP image embeddings and using them for classification tasks is a common practice, and serves as the baseline approach in this work.

The rest of the manuscript is organized as follows. Section \ref{sec:method} presents in detail the proposed method for disturbing image detection exploiting LMM-knowledge. Subsequently, Section \ref{sec:exp} provides the experimental evaluation of the proposed method, followed by the conclusions in Section \ref{sec:con}.

%
%
\section{Proposed Method}\label{sec:method}
\begin{figure*}
\centerline{\includegraphics[width=\textwidth]{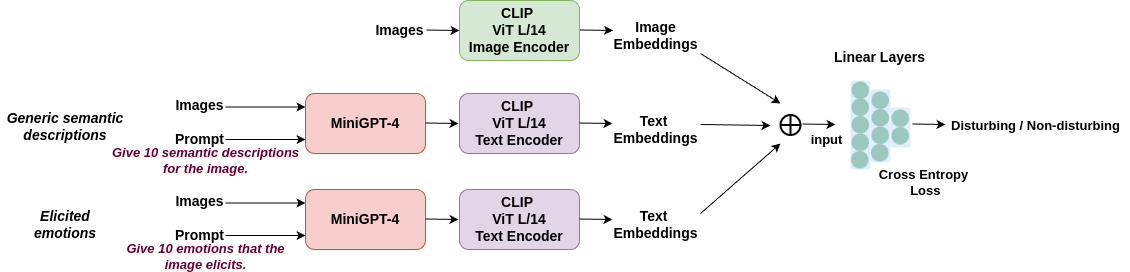}}
\caption{Proposed method for \textit{Disturbing Image Detection}. We first prompt the MiniGPT-4 model for obtaining 10 generic semantic descriptions for each image of the dataset. We also prompt the MiniGPT-4 model for obtaining 10 elicited emotions for each image of the dataset. Then we extract the CLIP embeddings for both the MiniGPT-4-generated responses. Finally, these two text embeddings are concatenated with the corresponding CLIP image embeddings and propagated to the linear layers for performing the DID task, using cross entropy loss.}
\label{fig:proposed-method}
\end{figure*}

In this section we present the proposed method for DID, briefly presenting first its two main components, i.e., MiniGPT-4 and CLIP.

\subsection{MiniGPT-4}
GPT-4 \cite{achiam2023gpt} is the first language model that accepts both text and image input, and generates text output. Since its technical details remain undisclosed, MiniGPT-4 was afterwards proposed. MiniGPT-4 aligns a frozen visual encoder with a frozen LLM, using a projection layer. Particularly, it uses the Vicuna LLM \cite{vicuna2023}, while for the visual perception it uses a ViT-G/14 \cite{zhai2022scaling} from EVA-CLIP \cite{sun2023eva} and a Q-Former network. 

\subsection{CLIP}
CLIP is the first VLM trained with natural language supervision at large scale, even though VLMs have been emerged since 2015 \cite{vinyals2015show}. It consists of an image and a text encoder, and it is trained with image-text pairs in order to predict which of the possible pairings actually occurred. 
To achieve this goal, it learns a multimodal embedding space by jointly training the image and text encoders to maximize the cosine similarity of the corresponding image and text embeddings of the correct pairs, while minimizing the cosine similarity of the embeddings of the incorrect pairings.  CLIP has been established as a powerful zero-shot classifier. Besides, a common approach to leverage CLIP for classification tasks is to use its image encoder for extracting the corresponding image embeddings and use them with a linear classifier, which also accomplishes exceptional performance. Hence, this approach serves as our baseline, and our goal is to advance the classification performance on the DID task, exploiting LMM-generated knowledge. 

\subsection{Exploiting LMM-generated responses for DID}
We propose to leverage LMM-encoded knowledge in a two-fold manner, i.e., by extracting generic semantic descriptions, as well as by extracting elicited emotions for each image of the dataset, as illustrated in Fig. \ref{fig:proposed-method}. To do so, we first prompt the MiniGPT-4 model with each image of the dataset along with the text \textquotedblleft \textit{Give 10 semantic descriptions for the image}\textquotedblright. Subsequently, we propagate the obtained responses to the CLIP's text encoder (i.e., 10 semantic descriptions for each image) in order to obtain the text embeddings at the final layer of the text encoder. Next, a pooling operation is applied on the produced text embeddings, resulting in a unique embedding for each image. This process can be applied to generic classification problems, as in e.g., \cite{eann2024}.

Moreover, we also propose a similar process that is tailored to the DID task. Specifically, we also prompt the MiniGPT-4 model with each image of the dataset along with the text \textquotedblleft \textit{Give 10 emotions that the image elicits}\textquotedblright. Correspondingly, we introduce the obtained responses to the CLIP's text encoder (i.e., 10 elicited emotions for each image) in order to obtain the corresponding text embeddings, followed again by a pooling operation that results in a unique embedding for each image. Then, we use both the semantic description and elicited emotion text embeddings along with the corresponding image embeddings, obtained from the image encoder of CLIP, to feed a classification head with a cross entropy loss for performing binary classification. 

More specifically, we consider a set $\mathcal{X} =\{\mathbf{X}_i \in \Re^{h\times w \times c} | i=1, \dots, N\}$ of $N$ images, where $h, w, c$ denote the height, width, and channels of the image, respectively. Each image $\mathbf{X}_i$ is associated with a class label $l_i \in \{0,1\}$. We use MiniGPT-4 first for obtaining generic semantic descriptions. That is, we consider for each image $\mathbf{X}_i$ the MiniGPT-4-generated semantic description responses, $s_i^j, j=1,\dots,10$, and then we extract the text embeddings using the CLIP's text encoder $\textit{g}$. That is, $\{\mathbf{t}_i^j=\textit{g}(s_i^j)  \in \Re^{D}| j=1,\dots,10\}$, where $D$ is the dimension of the embeddings. Subsequently, the average pooling operation is applied and a unique text embedding $\mathbf{t}^M_i  \in \Re^{D}$ is obtained for each image $\mathbf{X}_i$. Correspondingly, we obtain a unique embedding of the elicited emotions for each image of the dataset, i.e., $\mathbf{z}^M_i  \in \Re^{D}$. Finally, we use concatenation for combining the image embeddings, $\mathbf{y}_i = \textit{f}(\mathbf{X}_i) \in \Re^{D}$, where $\textit{f}$ corresponds to the CLIP's image encoder, with the text embeddings and performing the classification task. That is, $[\mathbf{y}_i^\intercal,\mathbf{t}^{M \intercal}_i,\mathbf{z}^{M \intercal}_i]^\intercal \in \mathbb{R}^{D+D+D}$. In the performed experiments we use CLIP with ViT-L-14, where the dimensions of both the image and text embeddings is 768, i.e., $D=768$. Then, a typical classification head consisting of three learnable linear layers processes the concatenated embedding to predict the class labels, using the cross entropy loss.

%
%
\section{Experimental Evaluation}\label{sec:exp}
In this section we present the experiments performed in order to evaluate the proposed method for DID. First, we present the utilized dataset, followed by the evaluation metrics and the implementation details. Finally, we provide the experimental results. 

\subsection{Dataset}
We use an augmented version of the DID dataset \cite{zampoglou2016web}, abbreviated as DID-Aug., developed in \cite{sarridis2022leveraging}. The dataset augments the DID training set using hard examples from the YFCC dataset \cite{thomee2016yfcc100m}. The DID-Aug. comprises of 30,106 training images, including 8,070 disturbing images and 22,036 non-disturbing images. The test set comprises of 1,080 images, including 405 disturbing images and 675 non-disturbing images. It should be noted that this is the largest dataset for DID, since as it is stated in \cite{sarridis2022leveraging}, collecting data for this problem constitutes a challenge task, due to the nature of the positive class.

\subsection{Evaluation Metrics}
Throughout this work, following the evaluation protocol used in \cite{sarridis2022leveraging}, we use test accuracy, i.e. the ratio of number of correct predictions to the total number of input samples, for evaluating the performance of the proposed method. Each experiment is executed five times and we report the mean value and the standard deviation, considering the maximum value of test accuracy for each experiment. We also provide qualitative results of the MiniGPT-4 responses. 

\subsection{Implementation Details}
We utilize MiniGPT-4 with Vicuna-13B locally for acquiring the generic semantic descriptions and elicited emotions, and the ViT-L-14 CLIP version for extracting the corresponding image and text embeddings. For the binary classification task of distinguishing between distubing and non-disturbing images we use three linear layers of 512, 256, and 2 neurons. The models are trained for 500 epochs, the learning rate is set to 0.001, and the batch size is set to 32 samples. Experiments are performed using the Pytorch framework on an NVIDIA GeForce RTX 3090 with 24 GB of GPU memory.

\subsection{Experimental Results}
In Table \ref{tab:res} we report the experimental results of the proposed method that uses CLIP text embeddings of elicited emotions and generic semantic descriptions along with the CLIP image embeddings against the baseline that uses only the CLIP image embeddings. Also, comparison against current state-of-the-art that uses an EfficientNet-b1 model is provided. Best performance is printed in bold. As can be observed, the proposed method significantly improves the baseline performance in terms of accuracy, accomplishing also superior performance over current state-of-the-art. 

\begin{table}
\centering
\caption{Test accuracy on DID-Aug. dataset - Comparison with state-of-the-art.} \label{tab:res}
\resizebox{0.46\textwidth}{!}{  
  \begin{tabular}{|c|c|}
  \hline
  \bf{Method} & \bf{Test accuracy (\%)} \\  \hline
  CLIP - Image Embeddings \cite{radford2021learning} & 94.444  \\ \hline
  CM-Refinery (EfficientNet-b1) \cite{sarridis2022leveraging} & 95.000 \\ \hline
  CLIP - Proposed & \bf{96.907} \\ \hline
\end{tabular}}
\end{table}

Subsequently, in Table \ref{tab:ablation} we provide an ablation study on the utilized dataset. That is,  we first evaluate the performance of each of the components individually. Test accuracy using only the CLIP image embeddings is initially presented, which serves as the baseline. Subsequently, test accuracy using the CLIP text embeddings of the LMM-elicited emotions is presented, followed by the corresponding results of the CLIP text embeddings of the LMM-generated semantic descriptions. Next, we provide all the combinations of the considered embeddings. 

\begin{table*}
\centering
\caption{Test accuracy on DID-Aug. dataset - Ablation study.} \label{tab:ablation}
  \begin{tabular}{|c|c|}
  \hline
  \bf{Method} & \bf{Test accuracy (\%)} \\  \hline
  CLIP - Image Embeddings & 94.444 $\pm$ 0.131 \\ \hline
  CLIP - Emotion Embeddings & 91.092 $\pm$ 0.108 \\ \hline
  CLIP - Semantic Description Embeddings & 92.592 $\pm$ 0.058\\ \hline
  CLIP - Image Embeddings + Emotion Embeddings  & 95.462 $\pm$ 0.101 \\ \hline
  CLIP - Image Embeddings + Semantic Description Embeddings & 96.222 $\pm$ 0.107 \\ \hline
  CLIP - Emotion Embeddings + Semantic Description Embeddings & 95.185 $\pm$ 0.261 \\ \hline
  CLIP - Image Embeddings + Emotion Embeddings + Semantic Description Embeddings (proposed) & \bf{96.907 $\pm$ 0.125} \\ \hline
\end{tabular}
\end{table*}

\begin{figure*}[!h]
\centering
\includegraphics[width=0.85\textwidth, height=0.17\textheight]{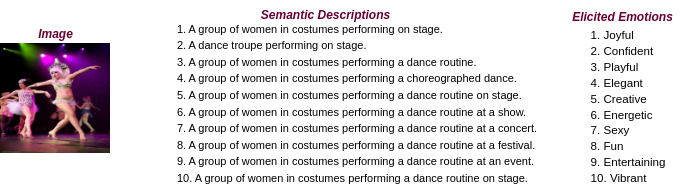}
\caption{Semantic descriptions and elicited emotions MiniGPT-4 responses for a non-disturbing image.}
\label{fig:qual-des-emotions-non}
\end{figure*}

\begin{figure*}[!h]
\centering
\includegraphics[width=0.85\textwidth, height=0.17\textheight]{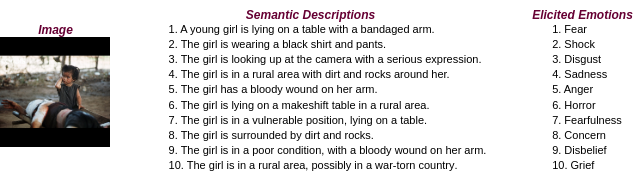}
\caption{Example of a test image that was misclassified by the baseline method, while correctly classified as \textit{disturbing} using the proposed method, along with the LMM-generated responses.}
\label{fig:misclassified}
\end{figure*}

From the demonstrated results several remarks can be drawn. First, it can be observed that the baseline approach provides very good performance, highlighting the  CLIP's image encoder capabilities. Next, we can observe that using only the MiniGPT-4 knowledge to represent the images of the dataset, results in, as expected, lower but very competitive performance. That is, using only the semantic description embeddings the model achieves accuracy 92.592\%. Interestingly, using only the LMM-elicited emotion embeddings the model achieves an accuracy of 91.092\%, while combining the two text embeddings we achieve a significant improvement, compared to using either one alone. Regarding the combination of image embeddings with each of the text embeddings, we can observe that this leads to enhanced performance. Finally, considering the proposed approach of using the image embeddings along with text embeddings of generic semantic descriptions and elicited emotions, we can observe it leads to the highest performance, validating our claim that we can accomplish significant improvements by incorporated knowledge encoded in LMMs. In particular, apart from the generic semantic description embeddings, the elicited emotion embeddings, which are tailored to the DID task, give further improvement.

Subsequently, in Fig. \ref{fig:qual-des-emotions-non} we provide some qualitative results of the knowledge encoded in the MiniGPT-4 model. More specifically, we present the responses of the model, prompted first with a non-disturbing image and the text \textquotedblleft \textit{Give 10 semantic descriptions for the image}\textquotedblright and next with the image and the text
\textquotedblleft \textit{Give 10 emotions that the image elicits}\textquotedblright. As it is demonstrated, the LMM provides meaningful semantic descriptions and emotions.  

Finally, in Fig. \ref{fig:misclassified} we present a disturbing test image that was  misclassified using the baseline method, while correctly classified using the proposed method, along with the LMM-generated responses. It is evident that the semantic descriptions provide useful information on the context, while the unpleasant elicited emotions (e.g., fear, sadness, grief) clearly assist the classifier towards the correct decision.

%
%
\section{Conclusions}\label{sec:con}
In this paper we dealt with the DID problem, leveraging knowledge encoded in LMMs. Specifically, we proposed to appropriately prompt the LMM in order to extract generic semantic descriptions, as well as elicited emotions. Then, we used the CLIP's text encoder in order to obtain the text embeddings of both the generic semantic descriptions and LMM-elicited emotions. Finally, we used them along with the corresponding CLIP's image embeddings for addressing the downstream task of DID. The proposed method accomplishes remarkable performance in terms of classification accuracy, achieving also superior performance over the current state-of-the-art on the DID-Aug. dataset.

%
%
\section*{Acknowledgment}
This work has been funded by the European Union as part of the Horizon Europe Framework Program, under grant agreements 101070190 (AI4TRUST) and 101070109 (TransMIXR).


\end{document}